\documentclass[11pt,a4paper]{article}
\usepackage[hyperref]{acl2019}
\usepackage{times}
\usepackage{latexsym}
\usepackage{amsmath}
\usepackage{graphicx}
\usepackage{tabularx}
\usepackage{float}
\usepackage[utf8]{inputenc}

\usepackage{url}

\aclfinalcopy 

\title{Towards Integration of Statistical Hypothesis Tests into Deep Neural Networks}

\author{Ahmad Aghaebrahimian \\
  Zurich University of Applied Sciences \\
  Switzerland \\
  \texttt{agha@zhaw.ch} \\\And
  Mark Cieliebak \\
  Zurich University of Applied Sciences \\
  Switzerland \\
  \texttt{ciel@zhaw.ch} \\}

\date{}

\begin{document}
\maketitle
\begin{abstract}
We report our ongoing work about a new deep architecture working in tandem with a statistical test procedure for jointly training texts and their label descriptions for multi-label and multi-class classification tasks. A statistical hypothesis testing method is used to extract the most informative words for each given class. These words are used as a class description for more label-aware text classification. Intuition is to help the model to concentrate on more informative words rather than more frequent ones. The model leverages the use of label descriptions in addition to the input text to enhance text classification performance. Our method is entirely data-driven, has no dependency on other sources of information than the training data, and is adaptable to different classification problems by providing appropriate training data without major hyper-parameter tuning. We trained and tested our system on several publicly available datasets, where we managed to improve the state-of-the-art on one set with a high margin, and to obtain competitive results on all other ones.
\end{abstract}

\section{Introduction}
Text classification is a complex problem in Natural Language Processing (NLP) with lots of applications from sentiment analysis~\cite{BingLiu} to question answering~\cite{aghaebrahimian-jurcicek-2016-open, Aghaebrahimian2016, yu:2014} or abusive language detection~\cite{Cieliebak, Antigoni-Founta-2018}, to name just a few. Text classification is defined as the task of assigning a certain pre-defined class to a document. 

The number of classes can be arbitrarily large in {\em multi-class} classification, whereas there are only two classes for binary classification. In {\em multi-label} classification, the number of labels attached to each document is not known and usually larger than one, while in {\em multi-class} classification, only one class is assigned to each document.
 
There exist numerous approaches for text classification, ranging from simple hand-crafted lexical-level features with Naive Bayes or Support Vector Machines (SVM)~\cite{S-Wang-2012} to self-learning approaches with Deep Neural Networks (DNN)~\cite{deriu}. 

For the latter, several architectures such as Convolutional or Recurrent Neural Networks (CNN or RNN)~\cite{Shen-2017, W-Wang-2018} have been proposed. These architectures learn different levels of textual representation in their layers, which are an essential source of information for the classification process. As an alternative, attention networks are also introduced~\cite{Bahdanau-2015, Yang-et-al-2016} to capture the features with the highest discriminative power regarding the class and irrespective of their distance. On the other hand, the field of Statistics has since long developed and optimized various methods to capture `relevant' properties of a given dataset. In this work, we extend DNNs with statistical hypothesis testing methods to enhance their performance in assessing feature relevancy on the input data.  

More precisely, our approach works as follows: 

- For each class, we generate a {\em class description}, which is a set of `most informative words' that will help to distinguish the class from others. 

- To achieve this, we apply two statistical hypothesis testing approaches called $\chi^2$ test~\cite{Pearson-1893} and Analysis of Variance test (ANOVA)~\cite{Fisher-1921}. 

- We then extend a DNN that is based on bidirectional Gated Recurrent Units (GRU) with an additional input channel for encoding the class descriptions. This channel uses attention, in addition, to enable the network to focus on the most informative words for each document and given each class.

Our experiments on four standard datasets show that this approach can already reach or even outperform state-of-the-art solutions for these datasets. While this is very promising, we want to stress already here that this is ongoing work, and that it needs extensive further experiments to fully understand when and why the proposed method works.

The main contributions of this work are the use of statistical hypothesis testing methods specifically for class descriptor extraction rather than feature extraction, and a new deep architecture working in tandem with a statistical test procedure with state-of-the-art performance in multi-label and multi-class classification.

We organize the remaining content into the following sections. After a review on state of the art in Section~\ref{RelatedWork}, we describe how we extract the class descriptors in Section~\ref{chi2}. Then we continue with a description of our deep architecture, followed by the system setup for our experiments in Sections~\ref{model} and~\ref{SystemSetup}, respectively. Finally we report our results in Section ~\ref{ExperimentalResults} and conclude in Section~\ref{Conclusion}.

\section{Related Work}
\label{RelatedWork}
 
Many solutions for text classification with DNNs use word embeddings as the primary and only representation of the input text. Word embeddings are low-dimensional word vectors that can be pre-computed using, for instance, Word2Vec~\cite{Mikolov-2013}, GloVe~\cite{Pennington-2014}, or fastText~\cite{fasttext}. Many studies show the importance of embeddings in text classification~\cite{Arora-2017, Wieting-2016}, and DNNs have been proven very useful in capturing syntactic- and semantic-level textual information. Still, it has been shown that they can benefit from additional representations of the input, e.g., by using a method called Attention Mechanism~\cite{Bahdanau-2015, Yang-et-al-2016, Cui-2016, J-Gehring-2017}. 

The main idea in the attention mechanism for text classification is to put emphasis on more informative tokens for each class. The attention mechanism has been successfully applied for different tasks in NLP, including but not limited to sentiment analysis~\cite{Zhou-2016}, modeling sentence pair~\cite{aghaebrahimian-2018-deep, Yin-2016}, question answering~\cite{aghaebrahimian-2018-linguistically, Seo-2016}, and summarization~\cite{Rush-2015}. 

The idea of joint learning of text and class descriptions is already practiced by~\citet{G-Wang-2018}. They showed that training an attention model on class descriptions in a joint embedding space is beneficial for text classification. However, they extracted class descriptions only from the class names, which limits the functionality of this approach, since in many use cases the names are very short, sometimes even just one token. 

In our work, the class descriptions are extracted using a statistical, data-driven approach. This makes our model independent from the label-set description which is not always available.

Our model is also similar to the one by~\citet{Antigoni-Founta-2018} with two differences. First, they assume that classes are provided with metadata, while our model extracts class descriptions directly from the training data. Second, the token channel and class description channel in our model both have the same time complexity thus they both converge simultaneously, and we do not need to worry about over-fitting one channel while the other is still training.

The use of statistical tests for extracting textual features in text classification is quite common and already proven beneficial.~\citet{BAHASSINE2018}, for instance, used three different statistical tests for feature extraction to improve Arabic text classification. However, we do not use statistical tests for feature extraction. Instead, we use them to extract class descriptions which are used as a second channel alongside with their accompanying texts in a deep neural network. 

This is the first time that statistical tests are used for extracting class descriptor tokens which can be used for jointly training deep neural models on texts with their class descriptions.

\section{Generating Class Descriptions}
\label{chi2}
We show how to extract class descriptions using a data-driven method applied to the training data. To retrieve the class descriptions, we use two statistical hypothesis testing approaches called $\chi^2$ and Analysis of Variance (ANOVA) tests. 

We assume that each class is triggered given specific tokens; hence, given each class, the frequencies of those tokens should be distinguishable from other non-triggering ones. Therefore, for each class, we formulate a {\em null hypothesis} (i.e., an indirect assumption) that states that the presence of specific tokens does {\em not} have any impact on their class determination. Then we check which words can reject the hypothesis, hence, have discriminative significance in distinguishing classes from each other.

The $\chi^2$ test is used to determine whether there is a significant difference between the expected frequencies and the observed frequencies of tokens in each one of the classes. Using the training data, we separate the documents into mutually exclusive classes. Given each class and the null hypothesis, we compute the $\chi^2$ of the documents' words, which provides us with the probability with which any token falls into the corresponding class. 

The $\chi^2$ test allows us to evaluate how likely it is to observe a word in a class, assuming the null hypothesis is true.

Similarly, we use the ANOVA F-statistics to determine whether there is a dependence between certain tokens and a specific class, so to eliminate the words that are most likely independent of the class, hence, irrelevant for classification.

Both tests provide each word with a probability given each class. To get an $n$-dimensional class description vector, we extract the $n$ top-rated words for each class and use them as {\em class descriptors}. One should be careful not to confuse word embedding dimensions with the dimension of class descriptions. By class description dimension, we mean the length of the string containing the most informative words given each class.

Some of the most informative words given the classes available in the AG News datasets~\cite{Corso-2005} are presented in Table~\ref{cho2words}.

\begin{table}[ht]
    \scalebox{0.81}{
    \begin{tabularx}{1.21\columnwidth}{|X|X|}
        \hline
        Class       & Informative words    \\
        \hline
        World       & iraq, minister, president, prime, baghdad, iraqi, dig, palestinian, military, nuclear, israeli, ... \\ 
        \hline
        Sports      & dig, season, league, team, game, cup, night, coach, victory, win, sports, championship, olympic, ... \\
        \hline
        Business    & oil, stocks, prices, percent, quickinfo, target, profit, company, shares, bilion, quarter, sales, earnings, ... \\
        \hline
        Science     & microsoft, software, internet, space, music, computer, users, web, search, windows, technology, ... \\ 
        \hline
    \end{tabularx}}
    \caption{Extracted $\chi^2$ words for the AG News dataset}
    \label{cho2words}
\end{table}

\section{Model}
\label{model}
\begin{figure}[t!]
  \includegraphics[width=\columnwidth]{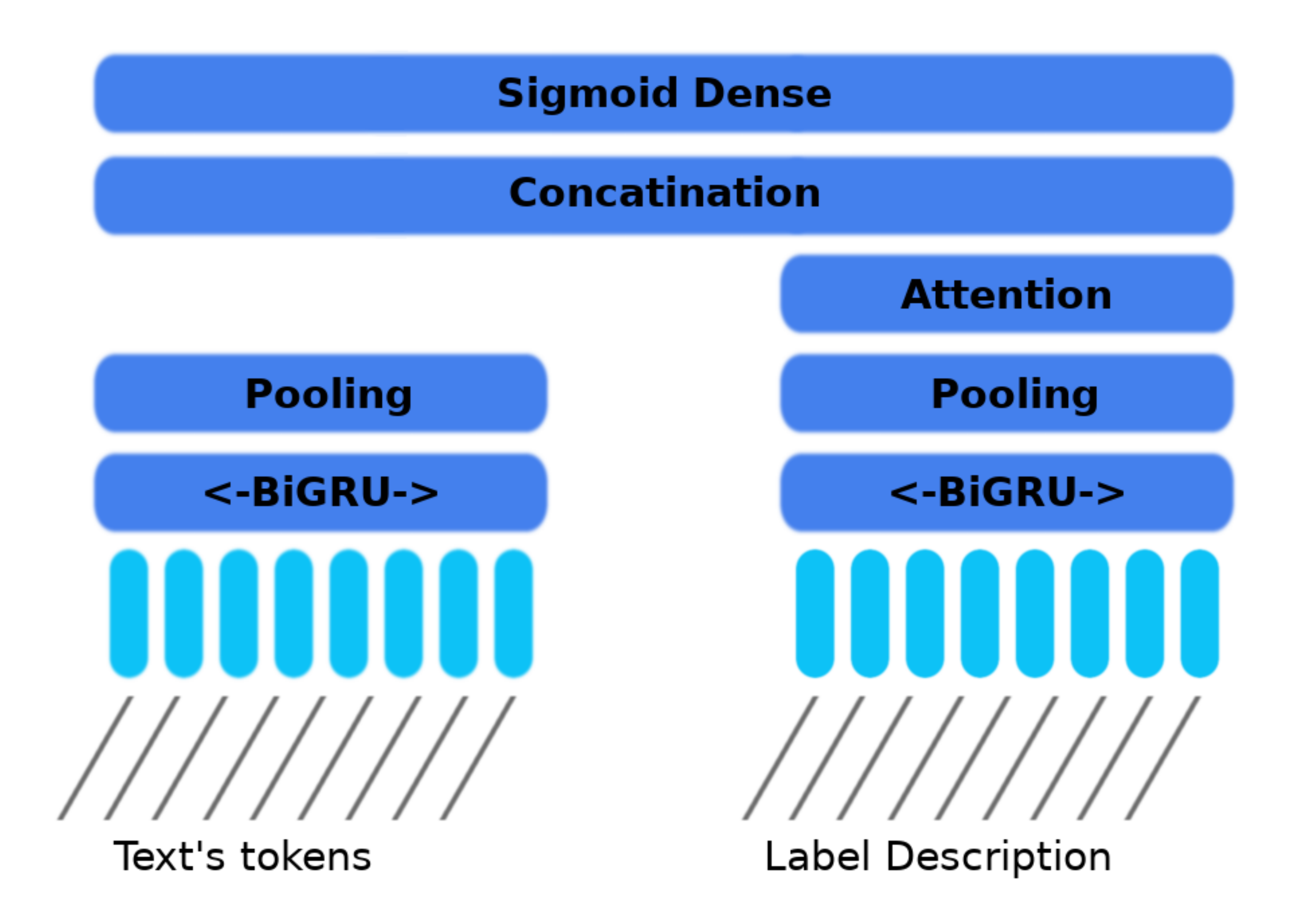}
  \caption{\label{arch}The system architecture}
\end{figure}

The overall architecture of our system is illustrated in Figure~\ref{arch}. We use a lookup table for transforming token indices to embedding vectors. The embeddings are fed into a bidirectional Gated Recurrent Unit (BiGRU)~\cite{Cho-2014}. The resulting tensors are then max- and average-pooled to extract the most relevant features. At the end of this channel, we have the vector of an encoded text. 
 
 A similar channel is used for encoding the class descriptions. Using the words extracted by the $\chi^2$ test as described in Section~\ref{chi2}, we generate a new string in which only the $\chi^2$ words are available. Given each class, this contains the highest informative words for this class.
 Additionally, we put an attention layer on top of this channel to learn the importance of each word given a particular class. 
 
 The attention layer is implemented based on the work of~\citet{Yang-et-al-2016}. The mathematical representation is as follow:
 
\begin{equation} \label{eq1}
\begin{split}
u & = f(\omega \cdot h + b) \\
a_i &  = Softmax(u_i \cdot u_s)\\
v_i &  = \sigma_i a_i \cdot h_i.
\end{split}
\end{equation}

where $h$ are the tensors out of the BiGRU layer, and $w$, $b$, $a$, and $v$ are the weight vectors, bias terms, attention vectors, and document vectors respectively.

Finally, we concatenate the resulting tensors from the attention layer with the max- and average-pooling layers and feed them into a dense layer for final classification.
 
For multi-class classification it is common to use the Softmax function \[P(c_j|x_i) = \frac{\exp(\omega_j \cdot x_i)}{\sum_{c=1}^C \exp(\omega_c \cdot x_i)}.\] where $x_i, c$, and $\omega$ are features, classes, and associated weight vectors, respectively. This function is used as the dense output layer of the network to normalize the logits generated by previous layers. In this way, we can model the probability of class $c_j$ as a multi-nominal distribution. The consequence of this decision is that the probability for a class is not independent of the other class probabilities, which would not be the desired behavior when dealing with a multi-label classification task. For instance, in a multi-label classification for hate speech detection, the probability of a comment for being offensive is independent of its probability of being hateful, because an offensive tone can be used in a text that is not necessarily hateful~\cite{Antigoni-Founta-2018}. For this reason, instead of Softmax, we use the Sigmoid activation function \[\sigma(z) = \frac{1}{1 + \exp(-z)}\] which is a better choice for multi-label classification. In this way we can model the probability of a class as Bernoulli's distribution \[P(c_j|x_i) = \frac{1}{1 + \exp(-\omega_j \cdot x_i)}\] which makes the probability of each class $c_j$ independent from the other class probabilities. 

Therefore we use a Softmax dense layer for multi-class and Sigmoid dense layer for multi-label classification to get the probabilities associated with each target class.

\section{System Setup}
\label{SystemSetup}
For the text representation in our system, we use pre-trained Glove embeddings~\cite{Pennington-2014} trained on 840 billion tokens with 300-dimensional vectors and set it to get updated through training. As the loss function for multi-class and multi-label settings, we use the Categorical and the Binary cross-entropy, respectively. We define the bidirectional GRUs, each with 128 units. We also set both the drop-outs~\cite{Srivastava:2014} and recurrent drop-outs to 0.5. In the following subsections, some details concerning the pre-processing of tested datasets are presented.

\subsection{Multi-label}
{\bf Data.} For the multi-label classification task, we train and test our model on a large publicly available dataset provided for toxic comment classification in a Kaggle competition called `Toxic Comment Classification Challenge.' The texts of the dataset were extracted from Wikipedia comments and have been labeled by human raters for six categories of toxic behavior: toxic, severe-toxic, obscene, threat, insult, and identity-hate. The training and test datasets contain 160k and 153k comments, respectively. The task is to train a model which assigns a probability to each of the six categories of toxic behavior given a new comment.

{\bf Pre-processing.} The pre-processing step for this dataset is performed by lower-casing, cleaning the comments from non-alphanumeric characters, using the first 130k most frequent tokens and removing comments longer than 80 tokens (95\% percentile of the training dataset).  Shorter comments are padded with zero to fixate the length of all comments to 80 tokens.

{\bf Performance Measure.} The Area Under the Receiver Operating Characteristic Curve (AUC-ROC) is used to measure the performance of the systems. ROC is a probability curve, and AUC is a measure of separability. This measure tells how much a model is capable of distinguishing between classes.

Since the output of the model is a vector of probabilities that the model computes for each class and we want to assign more than one class to each text, we define a threshold using the validation data and accept all the classes with probabilities above the threshold as positive class.

\subsection{Multi-class}
{\bf Data.} We also train and test our system on three other datasets for multi-class classifications, namely Hate Speech dataset~\cite{Davidson-2017}, AG News~\cite{Corso-2005}, and DBpedia, to measure its performance on multi-class classification. Some statistics of these datasets are reported in Table~\ref{ds:stats}. 

\begin{table}[H]
\scalebox{0.62}{
\begin{tabular}{|l|c|c|c|c|}
\hline
                                
Dataset                         &Type           &Classes/Labels         &Training       &Testing     \\ \hline
Hate Speech                     &Multi-class    &3                      &22.5K          &2.5K          \\ \hline
DBpedia                         &Multi-class    &14                     &560K           &70K          \\ \hline
AG News                         &Multi-class    &4                      &120K           &7.6K          \\ \hline
Kaggle-toxic comments           &Multi-label    &6                      &160K           &153K          \\ \hline
\end{tabular}}
\caption{Types, number of classes, and number of training/testing samples in the datasets used for training in this work}
\label{ds:stats}
\end{table}

{\bf Pre-processing.} The pre-processing step for these datasets is performed by lower-casing, removing non-alphanumeric characters, and removing repetitive characters from tokens (e.g. yoooouuuuu -\textgreater you).

{\bf Performance Measure.} In contrast to the multi-label setting, in the multi-class setting, we do not need to define a threshold. Instead, we get the argmax of the vector of probabilities since we need to return only one class.

\section{Experimental Results}
\label{ExperimentalResults}

Table~\ref{res1} shows that the system obtains superior results in the Hate Speech dataset and yields competitive results on the Kaggle data in comparison to some sate-of-the-art baseline systems. Table~\ref{res2} shows the results of our system on the DBpedia and AG News datasets. Using the same model without any tuning, we managed to obtain competitive results again compared to previous state-of-the-art systems. 

We also ran preliminary experiments on class description vectors with different dimensions (50 vs. 100), indicated by the suffix of each name in Table~\ref{res1}. By dimension, we mean the number of words given each label and not the dimension of word vectors which are all the same for both channels (i.e., 300). 

It turns out that in all but one case, the more words, the better the performance. However, we did not get statistically significant results with class descriptors with dimensions higher than 100. It seems that the range 50-100 is the optimal dimension for this approach and these datasets. Bigger vectors such as 150 did not yield any statistically significant improvement in performance, and 200-, and 300-dimensional vectors deteriorated the performance. 

We observed that the decline in the performance comes mainly from two sources: the network over-fit, and the similar words in different classes. By increasing the number of informative words, the number of similar words in different classes increases which leads to sub-optimal classification decision boundaries.

\begin{table}[ht]
\scalebox{0.75}{
\begin{tabular}{|l|c|c|c|c|}
\hline
\textbf{Hate Speech dataset}           &P(\%)              &R(\%)              &F1(\%)             &AUC(\%)            \\ \hline
\cite{Davidson-2017}            &91                 &90                 &90                 &87                 \\ \hline
\cite{Antigoni-Founta-2018}     &89                 &89                 &89                 &92                 \\ \hline
This work+$\chi^2$50            &89.7               &90.4               &90                 &92.9               \\ \hline
This work+$\chi^2$100           &90.3               &\textbf{92.5}      &\textbf{91.3}      &\textbf{93.7}      \\ \hline
This work+ANOVA50              &89.2               &89.6               &89.3               &92.1                \\ \hline
This work+ANOVA100             &89.8               &89.2               &89.4               &92.4                \\ \hline
\textbf{Kaggle dataset}         &                   &                   &                   &                   \\ \hline
Leader-board                    &-                  &-                  &-                  &98.82              \\ \hline
This work+$\chi^2$50            &-                  &-                  &-                  &98.05                \\ \hline
This work+$\chi^2$100           &-                  &-                  &-                  &98.24                \\ \hline
\end{tabular}}
\caption{The results of our system on the Hate Speech and Kaggle datasets. With one exception, in all cases longer class description leads to better performance. The results of the Kaggle dataset are only reported in AUC to be comparable with other systems in the multi-label category.}
\label{res1}
\end{table}

\begin{table}[ht]
\scalebox{0.75}{
\begin{tabular}{|l|c|c|c|c|}
\hline
                                &DBpedia(\%)            &AG News(\%)     \\ \hline
Bi-BloSAN\cite{Shen-2018}       &98.77              &93.32             \\ \hline
LEAM\cite{G-Wang-2018}          &99.02              &92.45             \\ \hline
This work                       &98.90              &92.05             \\ \hline
\end{tabular}}
\caption{Competitive results on DBpedia and AG News reported in accuracy (\%) without any hyper-parameter tuning.}
\label{res2}
\end{table}

\section{Conclusion}
\label{Conclusion}
Previous studies in text classification have shown that training classifiers with class descriptions or class metadata alongside the text is beneficial. However, many of these systems depend on the provided label set for generating their class descriptors. This dependence on an external source of information limits their applicability when such information is not available. 

In this paper, we proposed a data-driven approach for extracting class descriptions for jointly training text with their class descriptors, based on pure statistical tests. Moreover, we designed a new deep neural architecture to make use of the output of this statistical approach for enhancing the performance of text classification by attending on the informative words of each class. Although we have shown that the approach works in principle, by achieving state-of-the-art results on four standard datasets, it needs to be further explored in order to understand \textit{why} it works. In particular, we need to understand why words extracted with $\chi^2$ yield better results compared to ANOVA, how many words should be extracted given a specific task, if other statistical tests might even improve the outcomes, etc. Once this understanding is achieved, this may lead us towards proposing better data-driven approaches for extracting class descriptions that will be beneficial in text classification.
 
\bibliography{acl2019}

\begin{thebibliography}{34}
\expandafter\ifx\csname natexlab\endcsname\relax\def\natexlab#1{#1}\fi

\bibitem[{Aghaebrahimian(2018{\natexlab{a}})}]{aghaebrahimian-2018-deep}
Ahmad Aghaebrahimian. 2018{\natexlab{a}}.
\newblock Deep neural networks at the service of multilingual parallel sentence
  extraction.
\newblock In \emph{Proceedings of the 27th International Conference on
  Computational Linguistics (CoLing)}, pages 1372--1383, Santa Fe, New Mexico,
  USA. Association for Computational Linguistics.

\bibitem[{Aghaebrahimian(2018{\natexlab{b}})}]{aghaebrahimian-2018-linguistically}
Ahmad Aghaebrahimian. 2018{\natexlab{b}}.
\newblock Linguistically-based deep unstructured question answering.
\newblock In \emph{Proceedings of the 22nd Conference on Computational Natural
  Language Learning (CoNLL)}, pages 433--443, Brussels, Belgium. Association
  for Computational Linguistics.

\bibitem[{Aghaebrahimian and
  Jur{\v{c}}{\'\i}{\v{c}}ek(2016{\natexlab{a}})}]{Aghaebrahimian2016}
Ahmad Aghaebrahimian and Filip Jur{\v{c}}{\'\i}{\v{c}}ek. 2016{\natexlab{a}}.
\newblock Constraint-based open-domain question answering using knowledge graph
  search.
\newblock In \emph{Proceedings of the 19th International Conference of Text,
  Speech, and Dialogue (TSD)}, volume 9924, pages 28--36.

\bibitem[{Aghaebrahimian and
  Jur{\v{c}}{\'\i}{\v{c}}ek(2016{\natexlab{b}})}]{aghaebrahimian-jurcicek-2016-open}
Ahmad Aghaebrahimian and Filip Jur{\v{c}}{\'\i}{\v{c}}ek. 2016{\natexlab{b}}.
\newblock Open-domain factoid question answering via knowledge graph search.
\newblock In \emph{Proceedings of the Conference of the North American Chapter
  of the Association for Computational Linguistics (NAACL)- the Workshop on
  Human-Computer Question Answering}, pages 22--28, San Diego, California.
  Association for Computational Linguistics.

\bibitem[{Arora et~al.(2017)Arora, Liang, and Ma}]{Arora-2017}
Sanjeev Arora, Yingyu Liang, and Tengyu Ma. 2017.
\newblock A simple but tough-to-beat baseline for sentence embeddings.
\newblock In \emph{Proceedings of the International Conference on Learning
  Representations (ICLR)}.

\bibitem[{Bahassine et~al.(2018)Bahassine, Madani, Al-Sarem, and
  Kissi}]{BAHASSINE2018}
Said Bahassine, Abdellah Madani, Mohammed Al-Sarem, and Mohamed Kissi. 2018.
\newblock Feature selection using an improved chi-square for arabic text
  classification.
\newblock \emph{Journal of King Saud University - Computer and Information
  Sciences}.

\bibitem[{Bahdanau et~al.(2015)Bahdanau, Cho, and Bengio}]{Bahdanau-2015}
Dzmitry Bahdanau, Kyunghyun Cho, and Yoshua Bengio. 2015.
\newblock Neural machine translation by jointly learning to align and
  translate.
\newblock In \emph{Proceedings of the International Conference on Learning
  Representations (ICLR)}.

\bibitem[{Bojanowski et~al.(2017)Bojanowski, Grave, Joulin, and
  Mikolov}]{fasttext}
Piotr Bojanowski, Edouard Grave, Armand Joulin, and Tomas Mikolov. 2017.
\newblock Enriching word vectors with subword information.
\newblock \emph{Transactions of the Association for Computational Linguistics},
  5:135--146.

\bibitem[{Cho et~al.(2014)Cho, van Merrienboer, Bahdanau, and
  Bengio}]{Cho-2014}
Kyunghyun Cho, Bart van Merrienboer, Dzmitry Bahdanau, and Yoshua Bengio. 2014.
\newblock On the properties of neural machine translation: Encoder-decoder
  approaches.
\newblock In \emph{Proceedings of the Eighth Workshop on Syntax, Semantics and
  Structure in Statistical Translation}.

\bibitem[{Cui et~al.(2017)Cui, Chen, Wei, Wang, Liu, and Hu}]{Cui-2016}
Yiming Cui, Zhipeng Chen, Si~Wei, Shijin Wang, Ting Liu, and Guoping Hu. 2017.
\newblock Attention-over-attention neural networks for reading comprehension.
\newblock In \emph{Proceedings of the 55th Annual Meeting of the Association
  for Computational Linguistics (Volume 1: Long Papers)}, pages 593--602,
  Vancouver, Canada. Association for Computational Linguistics.

\bibitem[{Davidson et~al.(2017)Davidson, Warmsley, Macy, and
  Weber}]{Davidson-2017}
Thomas Davidson, Dana Warmsley, Michael~W. Macy, and Ingmar Weber. 2017.
\newblock Automated hate speech detection and the problem of offensive
  language.
\newblock In \emph{Proceedings of the International AAAI Conference on Web and
  Social Media (ICWSM)}.

\bibitem[{Del~Corso et~al.(2005)Del~Corso, Gull\'{\i}, and Romani}]{Corso-2005}
Gianna~M. Del~Corso, Antonio Gull\'{\i}, and Francesco Romani. 2005.
\newblock Ranking a stream of news.
\newblock In \emph{Proceedings of the 14th International Conference on World
  Wide Web}, WWW '05, pages 97--106, New York, NY, USA. ACM.

\bibitem[{Deriu and Cieliebak(2017)}]{deriu}
Jan~Milan Deriu and Mark Cieliebak. 2017.
\newblock Swissalps at semeval-2017 task 3 : attention-based convolutional
  neural network for community question answering.
\newblock In \emph{Proceedings of the 11th International Workshop on Semantic
  Evaluation (SemEval-2017)}, pages 334--338, Canberra, Australia. Association
  for Computational Linguistics.

\bibitem[{Fisher(1921)}]{Fisher-1921}
Ronald~A. Fisher. 1921.
\newblock On the `probable error' of a coefficient of correlation deduced from
  a small sample.
\newblock \emph{Metron}, 1:3--32.

\bibitem[{Founta et~al.(2018)Founta, Chatzakou, Kourtellis, Blackburn, Vakali,
  and Leontiadis}]{Antigoni-Founta-2018}
Antigoni-Maria Founta, Despoina Chatzakou, Nicolas Kourtellis, Jeremy
  Blackburn, Athena Vakali, and Ilias Leontiadis. 2018.
\newblock A unified deep learning architecture for abuse detection.
\newblock \emph{Computing Research Repository}, CoRR, abs/1802.00385.

\bibitem[{Gehring et~al.(2017)Gehring, Auli, Grangier, Yarats, and
  Dauphin}]{J-Gehring-2017}
Jonas Gehring, Michael Auli, David Grangier, Denis Yarats, and Yann~N Dauphin.
  2017.
\newblock Convolutional sequence to sequence learning.
\newblock \emph{arXiv preprint}, arXiv:1705.03122.

\bibitem[{von Grünigen et~al.(2018)von Grünigen, Grubenmann, Benites,
  Von~Däniken, and Cieliebak}]{Cieliebak}
Dirk von Grünigen, Ralf Grubenmann, Fernando Benites, Pius Von~Däniken, and
  Mark Cieliebak. 2018.
\newblock sp{MMMP} at {G}erm{E}val 2018 shared task: Classification of
  offensive content in tweets using convolutional neural networks and gated
  recurrent units.
\newblock In \emph{Proceedings of the GermEval 2018 Workshop : 14th Conference
  on Natural Language Processing (KONVENS)}.

\bibitem[{Liu(2015)}]{BingLiu}
Bing Liu. 2015.
\newblock \emph{Sentiment Analysis}.
\newblock Cambridge University Press, Cambridge, UK.

\bibitem[{Mikolov et~al.(2013)Mikolov, Chen, Corrado, and Dean}]{Mikolov-2013}
Tomas Mikolov, Kai Chen, Greg Corrado, and Jeffrey Dean. 2013.
\newblock Efficient estimation of word representations in vector space.
\newblock \emph{arXiv:1301.3781}.

\bibitem[{Pennington et~al.(1893)Pennington, Socher, and
  Manning}]{Pearson-1893}
Jeffrey Pennington, Richard Socher, and Christopher Manning. 1893.
\newblock Contributions to the mathematical theory of evolution.
\newblock In \emph{Proceedings of the Royal Society}.

\bibitem[{Pennington et~al.(2014)Pennington, Socher, and
  Manning}]{Pennington-2014}
Jeffrey Pennington, Richard Socher, and Christopher Manning. 2014.
\newblock Glove: Global vectors for word representation.
\newblock In \emph{Proceedings of the Conference on Empirical Methods in
  Natural Language Processing (EMNLP)}.

\bibitem[{Rush et~al.(2015)Rush, Chopra, and Weston}]{Rush-2015}
Alexander~M. Rush, Sumit Chopra, and Jason Weston. 2015.
\newblock A neural attention model for abstractive sentence summarization.
\newblock In \emph{Proceedings of the 2015 Conference on Empirical Methods in
  Natural Language Processing}, pages 379--389, Lisbon, Portugal. Association
  for Computational Linguistics.

\bibitem[{Seo et~al.(2016)Seo, Kembhavi, Farhadi, and Hajishirzi}]{Seo-2016}
Minjoon Seo, Aniruddha Kembhavi, Ali Farhadi, and Hannaneh Hajishirzi. 2016.
\newblock Bidirectional attention flow for machine comprehension.
\newblock \emph{arXiv:1611.01603}.

\bibitem[{Shen et~al.(2017)Shen, Zhang, Henao, Su, and Carin}]{Shen-2017}
Dinghan Shen, Yizhe Zhang, Ricardo Henao, Qinliang Su, and Lawrence Carin.
  2017.
\newblock Deconvolutional latent-variable model for text sequence matching.
\newblock \emph{CoRR}, abs/1709.07109.

\bibitem[{Shen et~al.(2018)Shen, Zhou, Long, Jiang, and Zhang}]{Shen-2018}
Tao Shen, Tianyi Zhou, Guodong Long, Jing Jiang, and Chengqi Zhang. 2018.
\newblock Bi-directional block self-attention for fast and memory-efficient
  sequence modeling.
\newblock In \emph{Proceedings of the International Conference on Learning
  Representations (ICLR)}.

\bibitem[{Srivastava et~al.(2014)Srivastava, Hinton, Krizhevsky, Sutskever, and
  Salakhutdinov}]{Srivastava:2014}
Nitish Srivastava, Geoffrey Hinton, Alex Krizhevsky, Ilya Sutskever, and Ruslan
  Salakhutdinov. 2014.
\newblock Dropout: A simple way to prevent neural networks from over fitting.
\newblock \emph{Journal of Machine Learning Research}, 15(1):1929--1958.

\bibitem[{Wang et~al.(2018{\natexlab{a}})Wang, Li, Wang, Zhang, Shen, Zhang,
  Henao, and Carin}]{G-Wang-2018}
Guoyin Wang, Chunyuan Li, Wenlin Wang, Yizhe Zhang, Dinghan Shen, Xinyuan
  Zhang, Ricardo Henao, and Lawrence Carin. 2018{\natexlab{a}}.
\newblock Joint embedding of words and labels for text classification.
\newblock In \emph{Proceedings of the Association for Computational Linguistics
  (ACL)}.

\bibitem[{Wang and Manning(2012)}]{S-Wang-2012}
Sida Wang and Christopher~D Manning. 2012.
\newblock Baselines and bigrams: Simple, good sentiment and topic
  classification.
\newblock In \emph{Proceedings of the Association for Computational Linguistics
  (ACL)}.

\bibitem[{Wang et~al.(2018{\natexlab{b}})Wang, Gan, Wang, Shen, Huang, Ping,
  Satheesh, and Carin}]{W-Wang-2018}
Wenlin Wang, Zhe Gan, Wenqi Wang, Dinghan Shen, Jiaji Huang, Wei Ping, Sanjeev
  Satheesh, and Lawrence Carin. 2018{\natexlab{b}}.
\newblock Topic compositional neural language model.
\newblock In \emph{Proceedings of the International Conference on Artificial
  Intelligence and Statistics (AISTATS)}.

\bibitem[{Wieting et~al.(2016)Wieting, Bansal, Gimpel, and
  Livescu}]{Wieting-2016}
John Wieting, Mohit Bansal, Kevin Gimpel, and Karen Livescu. 2016.
\newblock Towards universal paraphrastic sentence embeddings.
\newblock In \emph{Proceedings of the International Conference on Learning
  Representations (ICLR)}.

\bibitem[{Yang et~al.(2016)Yang, Yang, Dyer, He, Smola, and
  Hovy}]{Yang-et-al-2016}
Zichao Yang, Diyi Yang, Chris Dyer, Xiaodong He, Alex Smola, and Eduard Hovy.
  2016.
\newblock Hierarchical attention networks for document classification.
\newblock In \emph{Proceedings of the Conference of the North American Chapter
  of the Association for Computational Linguistics (NAACL)}.

\bibitem[{Yin et~al.(2016)Yin, Schutze, Xiang, and Zhou}]{Yin-2016}
Wenpeng Yin, Hinrich Schutze, Bing Xiang, and Bowen Zhou. 2016.
\newblock Abcnn: Attention-based convolutional neural network for modeling
  sentence pairs.
\newblock \emph{Transactions of the Association for Computational Linguistics
  (TACL)}.

\bibitem[{Yu et~al.(2014)Yu, Moritz~Hermann, Blunsom, and Pulman}]{yu:2014}
Lei Yu, Karl Moritz~Hermann, Phil Blunsom, and Stephen Pulman. 2014.
\newblock Deep learning for answer sentence selection.
\newblock In \emph{Proceedings of the Conference on Neural Information
  Processing Systems (NIPS) - Deep learning workshop}.

\bibitem[{Zhou et~al.(2016)Zhou, Wan, , and Xiao}]{Zhou-2016}
Xinjie Zhou, Xiaojun Wan, , and Jianguo Xiao. 2016.
\newblock Attention-based lstm network for cross-lingual sentiment
  classification.
\newblock In \emph{Proceedings of the conference on Empirical Methods in
  Natural Language Processing (EMNLP)}.

\end{thebibliography}
\bibliographystyle{acl_natbib}

\end{document}